\documentclass[draftclsnofoot, onecolumn]{ieeeaccess}
\usepackage{cite}
\usepackage{amsmath,amssymb,amsfonts}
\usepackage{algorithmic}
\usepackage{graphicx}
\usepackage{textcomp}
\usepackage{array}
\usepackage{booktabs}
\usepackage{color}
\usepackage{caption}
\usepackage{subcaption}
\newcolumntype{P}[1]{>{\centering\arraybackslash}p{#1}}
\newcolumntype{M}[1]{>{\centering\arraybackslash}m{#1}}

\def\BibTeX{{\rm B\kern-.05em{\sc i\kern-.025em b}\kern-.08em
    T\kern-.1667em\lower.7ex\hbox{E}\kern-.125emX}}
\begin{document}
\history{Preprint submitted to IEEE}
\doi{}
\title{Robust Segmentation Models using an Uncertainty Slice Sampling Based Annotation Workflow}

\author{\uppercase{Grzegorz Chlebus}\authorrefmark{1,2},
\uppercase{Andrea Schenk\authorrefmark{1}, 
Horst K. Hahn\authorrefmark{1,3},\\
Bram van Ginneken\authorrefmark{1,2},
and Hans Meine}\authorrefmark{1,4}}
\address[1]{Fraunhofer Institute for Digital Medicine MEVIS, 28359 Bremen, Germany}
\address[2]{Diagnostic Image Analysis Group, Department of Medical Imaging, Radboud University Medical Center, 6525 Nijmegen, The Netherlands}
\address[3]{Jacobs University, 28759 Bremen, Germany}
\address[4]{University of Bremen, Medical Image Computing Group, 28359 Bremen, Germany}

\markboth
{Chlebus G. \headeretal: Robust Segmentation Models via Uncertainty Slice Sampling Based Annotation Workflow}
{Chlebus G. \headeretal: Robust Segmentation Models via Uncertainty Slice Sampling Based Annotation Workflow}

\corresp{Corresponding author: Grzegorz Chlebus (e-mail: grzegorz.chlebus@mevis.fraunhofer.de).}

\begin{abstract}
Semantic segmentation neural networks require pixel-level annotations in large quantities to achieve a good performance.
In the medical domain, such annotations are expensive, because they are time-consuming and require expert knowledge.
Active learning optimizes the annotation effort by devising strategies to select cases for labeling that are most informative to the model.

In this work, we propose an uncertainty slice sampling (USS) strategy for semantic segmentation of 3D medical volumes that selects 2D image slices for annotation and compare it with various other strategies. 
We demonstrate the efficiency of USS on a CT liver segmentation task using multi-site data.
After five iterations, the training data resulting from USS consisted of 2410 slices (4\% of all slices in the data pool) compared to 8121\,(13\%), 8641\,(14\%), and 3730\,(6\%) for uncertainty volume (UVS), random volume (RVS), and random slice (RSS) sampling, respectively.

Despite being trained on the smallest amount of data, the model based on the USS strategy evaluated on 234 test volumes significantly outperformed models trained according to other strategies and achieved a mean Dice index of 0.964, a relative volume error of 4.2\%, a mean surface distance of 1.35\,mm, and a Hausdorff distance of 23.4\,mm.
This was only slightly inferior to 0.967, 3.8\%, 1.18\,mm, and 22.9\,mm achieved by a model trained on all available data, but the robustness analysis using the 5\textsuperscript{th} percentile of Dice and the  95\textsuperscript{th} percentile of the remaining metrics demonstrated that USS resulted not only in the most robust model compared to other sampling schemes, but also outperformed the model trained on all data according to Dice (0.946 vs. 0.945) and mean surface distance (1.92\,mm vs. 2.03\,mm).

\end{abstract}


\titlepgskip=-15pt

\maketitle

\section{Introduction}
\label{sec:introduction}
\PARstart{S}{emantic} segmentation of medical images plays a key role in many treatment planning workflows. 
Over the recent years, segmentation algorithms utilizing deep neural networks have provided state-of-the-art results for many segmentation tasks\cite{antonelli2021medical, bogunovic2019retouch, bilic2019liver}. 
Training of such systems typically requires large data sets with pixel-level annotations to achieve good performance. 
In the medical domain, such annotations require expert knowledge and are time-consuming, and thus, are expensive to obtain.
Although many pre-trained segmentation models are available, it is known that neural networks underperform when applied to data coming from different sites or imaging protocols\cite{gibson2018inter, karani2018lifelong}.
Therefore, to get the optimal performance, training on annotated target data is recommended. 

Active learning is a prominent technique for the optimization of annotation effort \cite{cohn1996active}.
It aims to design an optimal way to select training data resulting in high performance with low annotation cost.
Numerous active learning algorithms are pool-based, meaning that they employ query strategies to choose samples from the unlabeled data pool.
Pool-based query strategies typically take the model's uncertainty\cite{wang2016cost, gorriz2017cost}, sample representativeness\cite{sener2017active}, or both of them\cite{yang2017suggestive, smailagic2018medal} into account.

Smailagic et al.~proposed a strategy for a classification problem that uses predictive entropy to identify most uncertain examples in the data pool\cite{smailagic2018medal}.
Among these examples, those that maximize the distance to the training set in a feature space defined by one of the model layers are selected.
Addition of the distance maximization criterion allowed for a 32\% reduction in the number of required labeled examples.
A similar approach for a segmentation task was demonstrated by Yang et al. \cite{yang2017suggestive}, where an ensemble was used to pick uncertain cases defined as ones with the highest average per-pixel variance.
Additionally, examples with the highest representativeness according to the cosine similarity distance based on descriptors obtained from the model were picked for manual annotation.
Models trained using this approach achieved a state-of-the-art performance using only 50\% of the training data.

Wang et al.~introduced cost-effective active learning for image classification where in addition to manually labeled uncertain cases, ones with high confidence are pseudo-labeled by the model and added to the training set with no annotation effort\cite{wang2016cost}. 
A variational adversarial query strategy was proposed by Sinha et al.~that employs a discriminator network to differentiate between labeled and unlabeled examples\cite{sinha2019variational}.
This approach chooses points that are not well represented in the labeled set without the need for uncertainty measurement on the main task.

While most of the pool-based active learning research focuses on uncertainty quantification and example representativeness, little attention was given to the incorporation of partial annotations into the workflow. 
Partial annotations are particularly relevant for semantic segmentation tasks as they can result in a substantial annotation effort reduction\cite{cciccek20163d}, especially when dealing with volumetric data. In this work, we propose an uncertainty slice sampling (USS) query strategy to optimize the annotation workflow of 3D medical images.
Our strategy selects 2D image slices for annotation from a pool of unlabeled 3D volumes using predictive entropy as the uncertainty measure.
We see this as a way to increase the variability of the training set without the need for explicit modeling of example representativeness.
We show the efficiency of our strategy on a CT liver segmentation task using multi-site data.
We designed our experiments in such a way that they use a comparable annotation effort. We analyze the proposed strategy together with several alternatives: uncertainty volume sampling (UVS), random volume sampling (RVS), and random slice sampling (RSS).


\section{Methods}
\label{sec:methods}

\subsection{Neural Network Architecture}
We use a 3D anisotropic u-net (au-net) architecture (see Fig.\,\ref{fig:aunet}) that is a modified version of the commonly used encoder-decoder u-net segmentation network design\,\cite{ronneberger2015u}.
The model works on five resolution levels.
In the upper two levels, convolutional layers work only along $x$ and $y$ spatial dimensions and the remaining levels contain separable 3D convolutions to minimize the number of trainable parameters.
Each convolution layer is followed by batch normalization\cite{ioffe2015batch} and ReLU.
Max pooling (transposed 3D conv) layers are used in the encoder (decoder) for transitioning down (up) between resolution levels.
Dropout layer\cite{srivastava2014dropout} with a drop rate $p=0.25$ is placed at each resolution level in the decoder path to prevent overfitting and facilitate Monte Carlo (MC) sampling used for the uncertainty estimation.

\Figure(topskip=0pt, botskip=0pt, midskip=0pt)[scale=0.36]{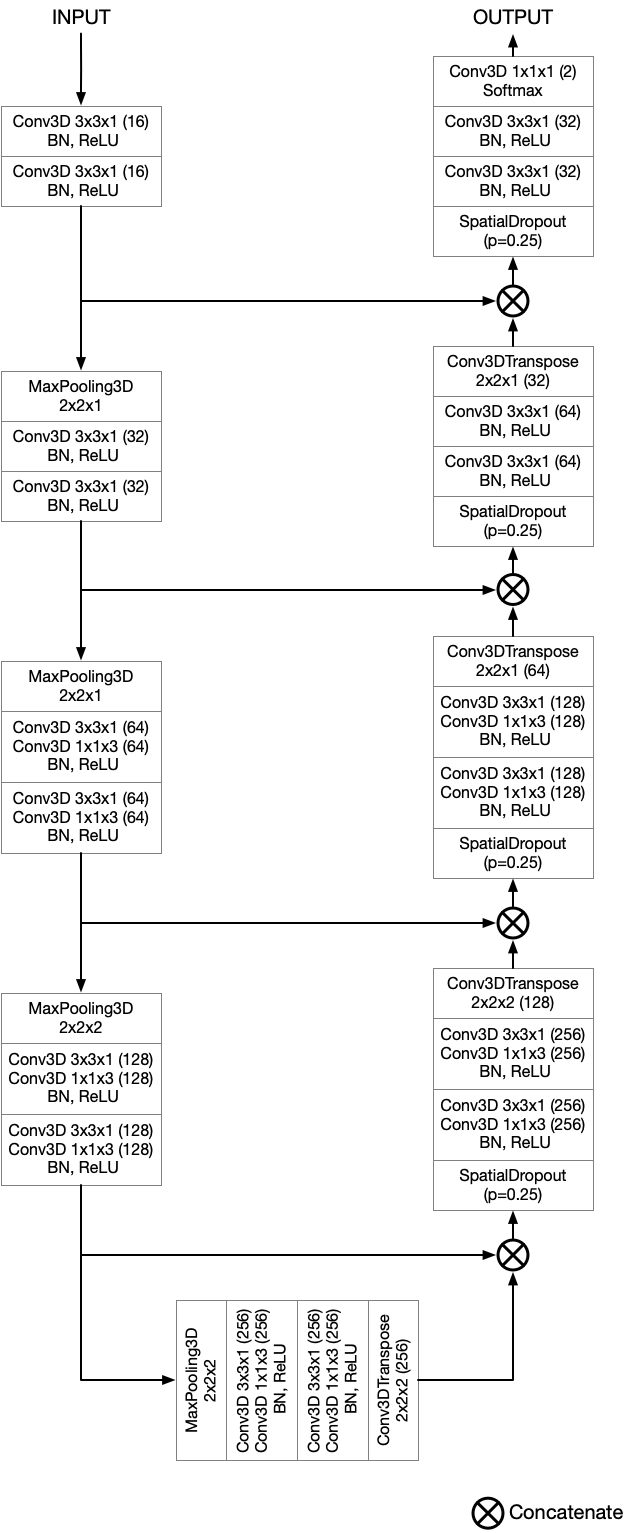}
{Anisotropic u-net (au-net) architecture. The model has 4 975 346 trainable parameters. \label{fig:aunet}}

\subsection{Training Setup}
All models are trained using a mini-batch size of 2 using $180\times180\times4$ image patches that are padded (reflect mode) on each side with $92$ voxels along $x$ and $y$ and $20$ along $z$ spatial dimension to account for valid convolutions.
Optimization is done using the Adam optimizer with $10^{-5}$ learning rate.
The model is applied to the validation data every 1000 iterations and the best model according to the Jaccard index is used for the final evaluation.

Stratified patch sampling is employed to speed up the training by ensuring that at least one patch in a mini-batch contains liver pixels.
We use a weighted soft dice loss to enable training with partially annotated patches (required for slice sampling strategies)\cite{sudre2017generalised}:

\begin{equation}
     L_{DSC} = 1 - \frac{2 \sum_{i}{w_i  y_i  p_i}}{\sum_i w_i y_i + \sum_i w_i p_i}
\label{eq:dice_loss}
\end{equation}
where $y_i$ is $i$th voxel label (1 - liver, 0 - background), $p_i$ is the output probability of the liver class, $w_i$ is 1 for annotated voxels and 0 otherwise, and $i$ runs over all voxels in a mini-batch.

\subsection{Data Preprocessing}
For training, all CT volumes were rescaled into Hounsfield units and resampled to $1.0\,\textrm{mm}\times1.0\,\textrm{mm}\times1.5\,\textrm{mm}$ voxel size.

\subsection{Uncertainty Estimation}
Several uncertainty estimation methods for segmentation models have been proposed including volume variation coefficient\cite{roy2018inherent}, prediction variance\cite{yang2017suggestive}, predictive entropy, and mutual information\cite{mukhoti2018evaluating}.
These methods require multiple samples, which are typically obtained from an ensemble or via MC dropout\cite{gal2016dropout}.
In our work, we use the predictive entropy as the uncertainty measure per voxel $\mathbf{x}$:

\begin{equation}
    U(\mathbf{x}) = \\ -\sum_{c}\left(\frac{1}{n}\sum_{n}p_n(y=c|\mathbf{x})\right)\mathrm{ln}\left(\frac{1}{n}\sum_{n}p_n(y=c|\mathbf{x})\right)
\label{eq:entropy}
\end{equation}
where $c$ is the number of classes, $n$ is the number of samples,
 $p_n(y=c|\mathbf{x})$ is the softmax probability of input $\mathbf{x}$ belonging to class $c$ in the $n$th sample. 
In our experiments, we obtain $n=20$ samples via MC dropout to capture the epistemic uncertainty accounting for the uncertainty in the model parameters\cite{kendall2017uncertainties}.

\noindent
\textbf{Volume-level uncertainty} Similarly to \cite{mehrtash2020confidence}, we use the average of the voxel-wise predictive entropy as a measure of the uncertainty at a volume-level. 
We exclude voxels outside of a dilated liver mask ($11\times11\times11$ kernel) for volume size invariance.

\noindent
\textbf{Slice-level uncertainty} is computed as the average of the predictive entropy for all voxels belonging to a given slice.
Contrary to the volume-level uncertainty, we compute the average over all voxels, because the slice-level measure is less sensitive to liver size variability.

\subsection{Query Strategies}
Query strategies select which samples from the unlabeled data pool should be annotated and added to the training set. 
In our study, we investigate two orthogonal dimensions of query strategies: i)~selection of 3D image volumes vs. 2D image slices and ii)~random vs. uncertainty-based sampling. These dimensions define four sampling strategies that work as follows:

\noindent
\textbf{Uncertainty Volume Sampling (UVS)}
selects volumes with the biggest volume-level uncertainty.

\noindent
\textbf{Random Volume Sampling (RVS)}
selects volumes at random.

\noindent
\textbf{Uncertainty Slice Sampling (USS)}
selects slice candidates with the biggest slice-level uncertainty. A slice candidate is extracted from a slice-level uncertainty vs. slice index curve using local maxima (see Fig.\,\ref{fig:slice_uncertainty_profile}). We require a minimum distance between peaks to be five slices to avoid selecting neighboring slices.

\noindent
\textbf{Random Slice Sampling (RSS)}
selects slices at random.

\Figure(topskip=0pt, botskip=0pt, midskip=0pt)[scale=0.6]{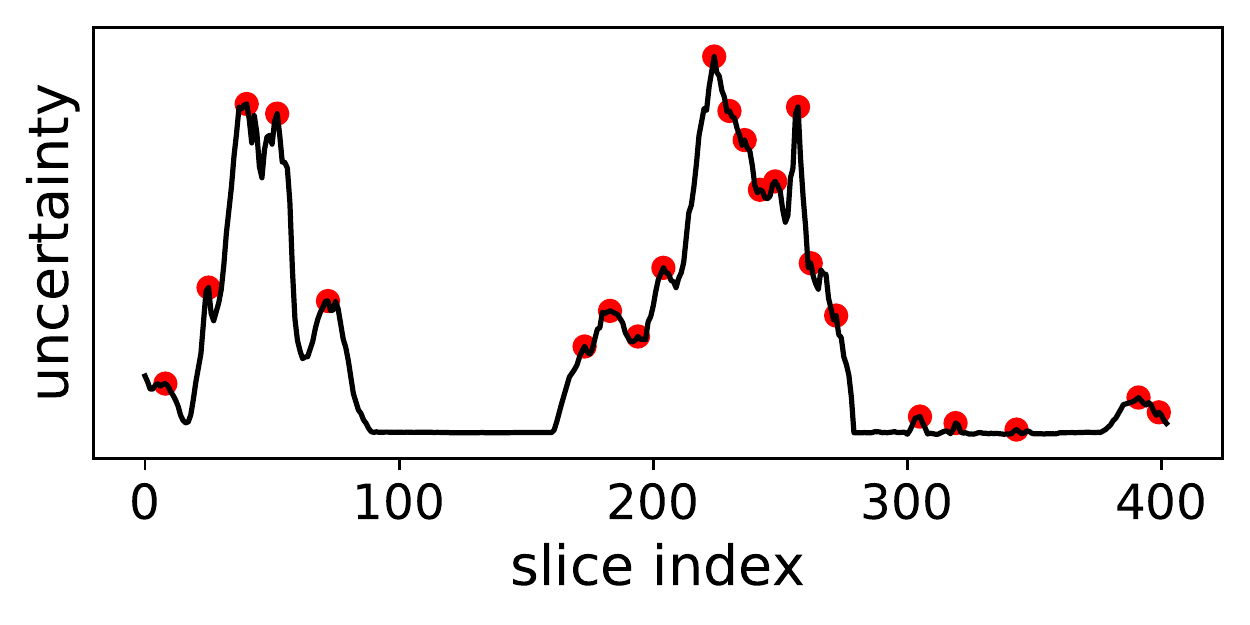}
{Slice uncertainty profile. The red dots correspond to slice candidates available to the uncertainty slice sampling strategy. \label{fig:slice_uncertainty_profile}}

\subsection{Evaluation Metrics}
We evaluate the segmentation quality with four commonly used metrics: Dice index (DICE),
relative volume error (RVE), mean surface distance (MSD), and Hausdorff distance (HD)\cite{chlebus2019reducing}. 
These metrics are defined as follows:
\begin{equation}
  \textrm{DICE}(X, Y) = \frac{2|X \cap Y|}{|X| + |Y|}
\end{equation}
\begin{equation}
  \textrm{RVE}(X, Y) = \frac{|V_X - V_Y|}{V_Y} \cdot 100\%
\end{equation}

\begin{equation}
  \textrm{MSD}(X, Y) = \frac{1}{2N} \Big( \sum_{x \in X} \min_{y \in Y} d(x, y) + \sum_{y \in
    Y} \min_{x \in X} d(x, y) \Big)
\end{equation}

\begin{equation}
  \textrm{HD}(X, Y) = max \{\sup_{x \in X} \inf_{y \in Y} d(x,y), \sup_{y \in Y} \inf_{x \in X} d(x,y) \}
\end{equation}
where $V_X$ and $V_Y$ are volumes of the test and the reference object, respectively, and $d(x,y)$ is the Euclidean distance between points $x$ and $y$.

In addition to reporting evaluation results on the test set using mean $\pm$ standard deviation, we report the 5\textsuperscript{th} percentile for the DICE and the 95\textsuperscript{th} percentile for the remaining metrics.
We use these percentile-based measures to assess the impact of investigated query strategies on models' robustness.

\section{Data}
\label{sec:data}
In this work, we simulate the active learning-based annotation workflow by using abdominal CT volumes with reference liver segmentations.
To increase the statistical power of our study, we decided to use imaging data coming from five datasets containing together 484 CT volumes.
Three of them are proprietary and were obtained via cooperation with clinical partners: Yokohama City University, Yokohama, Japan (\textit{Yokohama}), St\"{a}dtisches Klinikum Dresden, Dresden, Germany (\textit{Dresden}), and Radboud University Clinical Center, Nijmegen, the Netherlands (\textit{Rumc}).
Two remaining two (\textit{LiTS}, \textit{CHAOS}) come from publicly organized challenges\cite{bilic2019liver, CHAOS2021}.
Reference liver segmentations for the proprietary data were created manually by experienced clinical experts using dedicated annotation software\cite{schenk2001local}.
The challenge data comes with training liver masks that we used as reference segmentations.
In the case of \textit{LiTS} data, 65 out of 131 volumes were excluded by our medical experts due to a poor segmentation quality.
The combined dataset was divided into three subsets containing 240, 10, and 234 cases that were used 
as data pool, validation set, and test set,  respectively.
The reader is referred to Tab.\,\ref{tab:data} for more details.

\begin{table*}
\centering
\caption{Dataset details.}
\label{tab:data}
\setlength{\tabcolsep}{3pt}
\begin{tabular}{l r r r}
\toprule
Dataset Name & \#Volumes (pool/val/test) & Voxel Size\,[mm]  & Resolution \\
\midrule
Yokohama & 218 (119/2/97) &  [0.5-0.8]$\times$[0.5-0.8]$\times$[0.8-1.6] & 512$\times$512$\times$[170-376] \\
Rumc & 100 (46/2/52) & [0.6-1.0]$\times$[0.6-1.0]$\times$[0.7-1.5] & 512$\times$512$\times$[378-1057] \\
Dresden & 80 (36/2/42) & [0.6-0.9]$\times$[0.6-0.9]$\times$[0.5-5.0] & 512$\times$512$\times$[71-1160] \\
LiTS & 66 (31/2/33) & [0.6-1.0]$\times$[0.6-1.0]$\times$[0.7-5.0] & 512$\times$512$\times$[74-987] \\ 
CHAOS & 20 (8/2/10) &  [0.6-0.8]$\times$[0.6-0.8]$\times$[1.0-2.0] & 512$\times$512$\times$[81-266] \\
\bottomrule
\end{tabular}
\end{table*}

\section{Experiments}
\label{sec:experiments}
Our goal was to compare the investigated query strategies by keeping the required annotation effort at a comparable level.
To this end, we made the following assumptions w.r.t. the annotation effort:
\begin{enumerate}
    \item the effort required for annotation of $N$ slices with liver in one image volume is approx. equal to the annotation effort of $N/3$ slices with liver coming from different volumes;
    \item the effort required for a slice without a liver is negligible in comparison to liver slices.
\end{enumerate}
The first assumption was derived empirically from an observation that for segmentation of one volume one needs to manually draw contours on average on every third slice when using modern segmentation tools with interpolation functionality.

\subsection{Query Strategy Comparison}
To evaluate the efficiency of the investigated query strategies, we performed five active learning iterations and compared models after each iteration. 
We used a model trained on five fully annotated cases as the initial model.
In each iteration, we trained a model from scratch and we used the best model according to the validation score for evaluation and uncertainty estimation. 
All  models were trained for a max of 30 epochs, which is short enough to guarantee that model training and selection of cases for annotation can be performed overnight.

For volume sampling strategies we added five volumes to the training set in each iteration and we denote by $\tilde{N}_S^{\textrm{liver}}$ the average count of liver slices added in each UVS iteration.
Based on our assumption w.r.t. the annotation effort, in the USS strategy we added $N_S = \tilde{N}_S^{\textrm{liver}}/3$ slices in each iteration, whereas in the RSS we sampled random slices until $N_S$ liver slices were selected.


\subsection{Converged Models Comparison}
To evaluate the representativeness of training sets obtained with the investigated query strategies after five active learning iterations, we compared models trained on these training sets until convergence.
This was motivated by the fact that, for some of the models, 30 epochs were not long enough to converge.

\begin{figure*}
     \centering
     \begin{subfigure}[b]{0.45\textwidth}
         \centering
         \includegraphics[width=\textwidth]{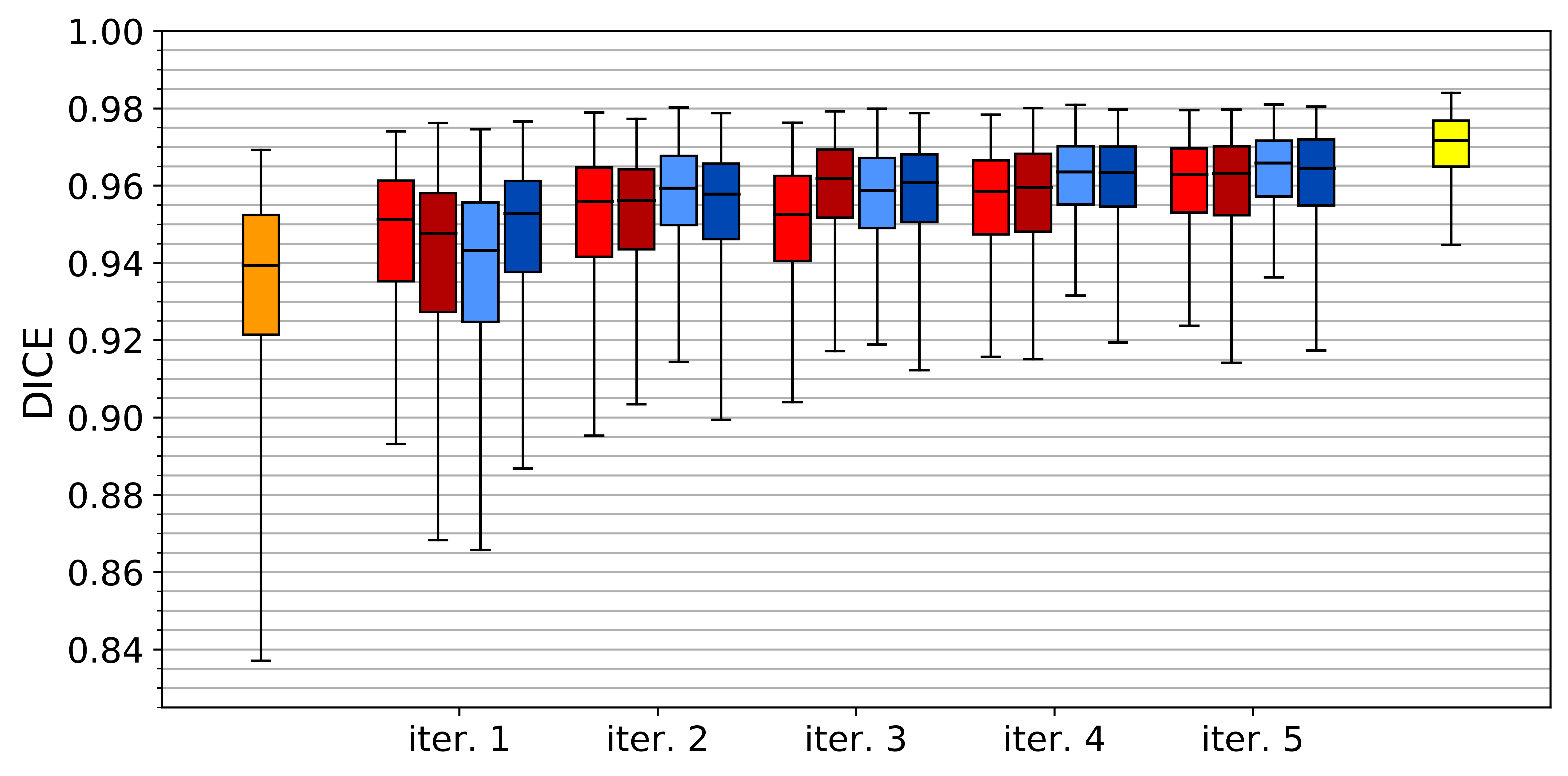}
         \caption{DICE (lower cap is 5\textsuperscript{th} percentile)}
         \label{fig:dice_over_iter}
     \end{subfigure}
     \hfill
     \begin{subfigure}[b]{0.45\textwidth}
         \centering
         \includegraphics[width=\textwidth]{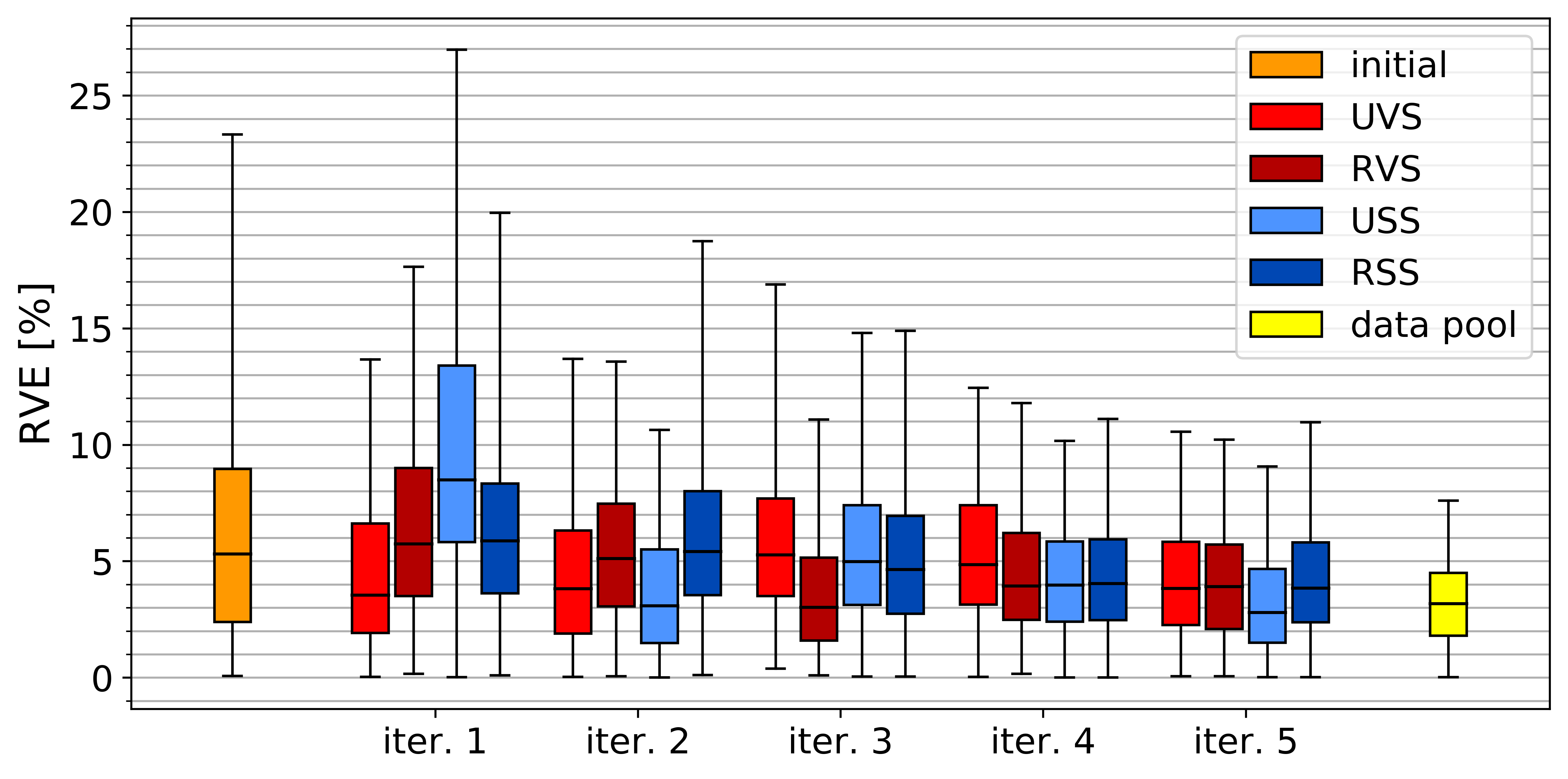}
         \caption{RVE (upper cap is 95\textsuperscript{th} percentile)}
         \label{fig:rve_iter}
     \end{subfigure}
    
    \vspace*{5mm}
     \begin{subfigure}[b]{0.45\textwidth}
         \centering
         \includegraphics[width=\textwidth]{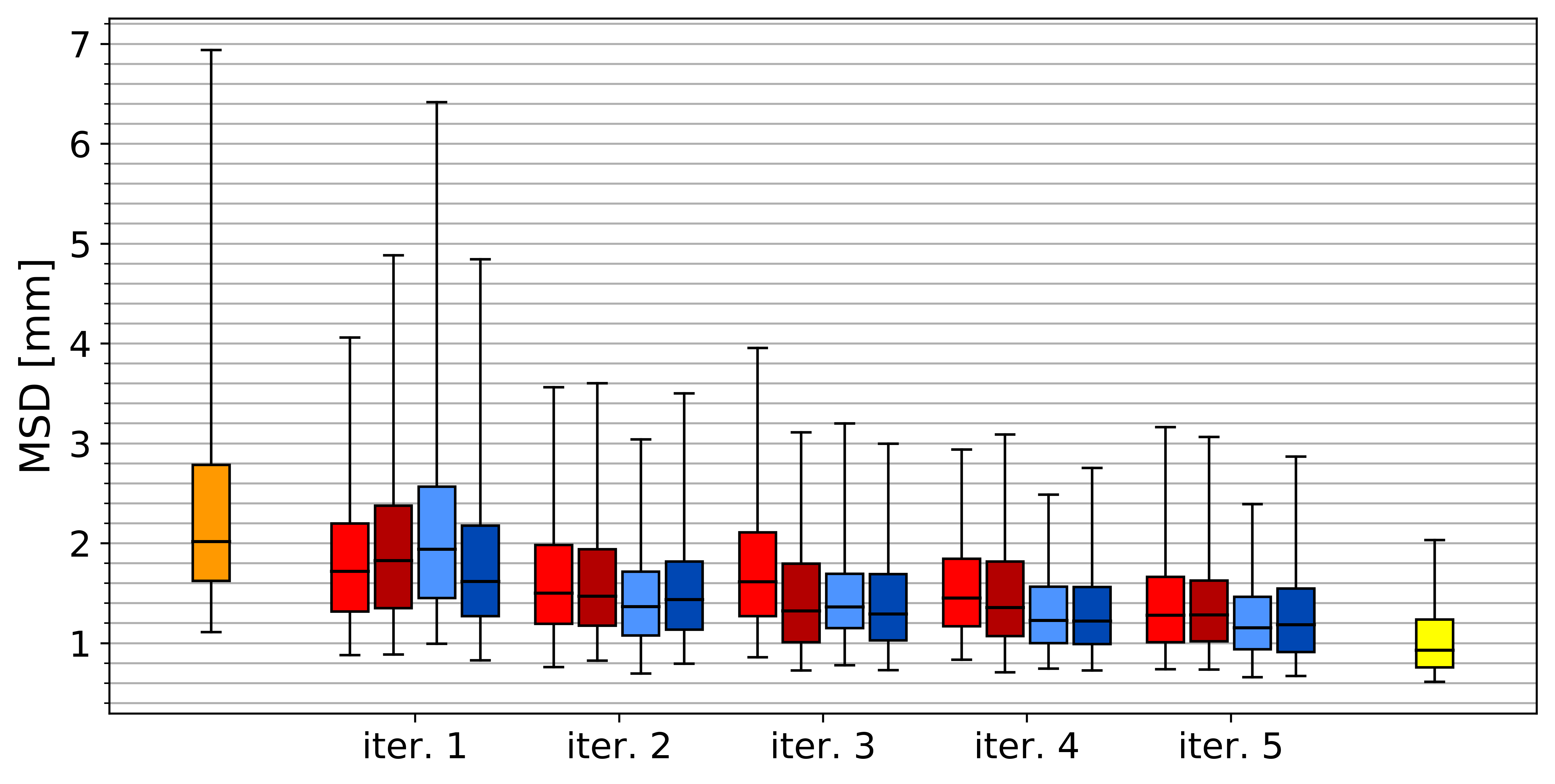}
         \caption{MSD (upper cap is 95\textsuperscript{th} percentile)}
         \label{fig:mean_dist_iter}
     \end{subfigure}
     \hfill
     \begin{subfigure}[b]{0.45\textwidth}
         \centering
         \includegraphics[width=\textwidth]{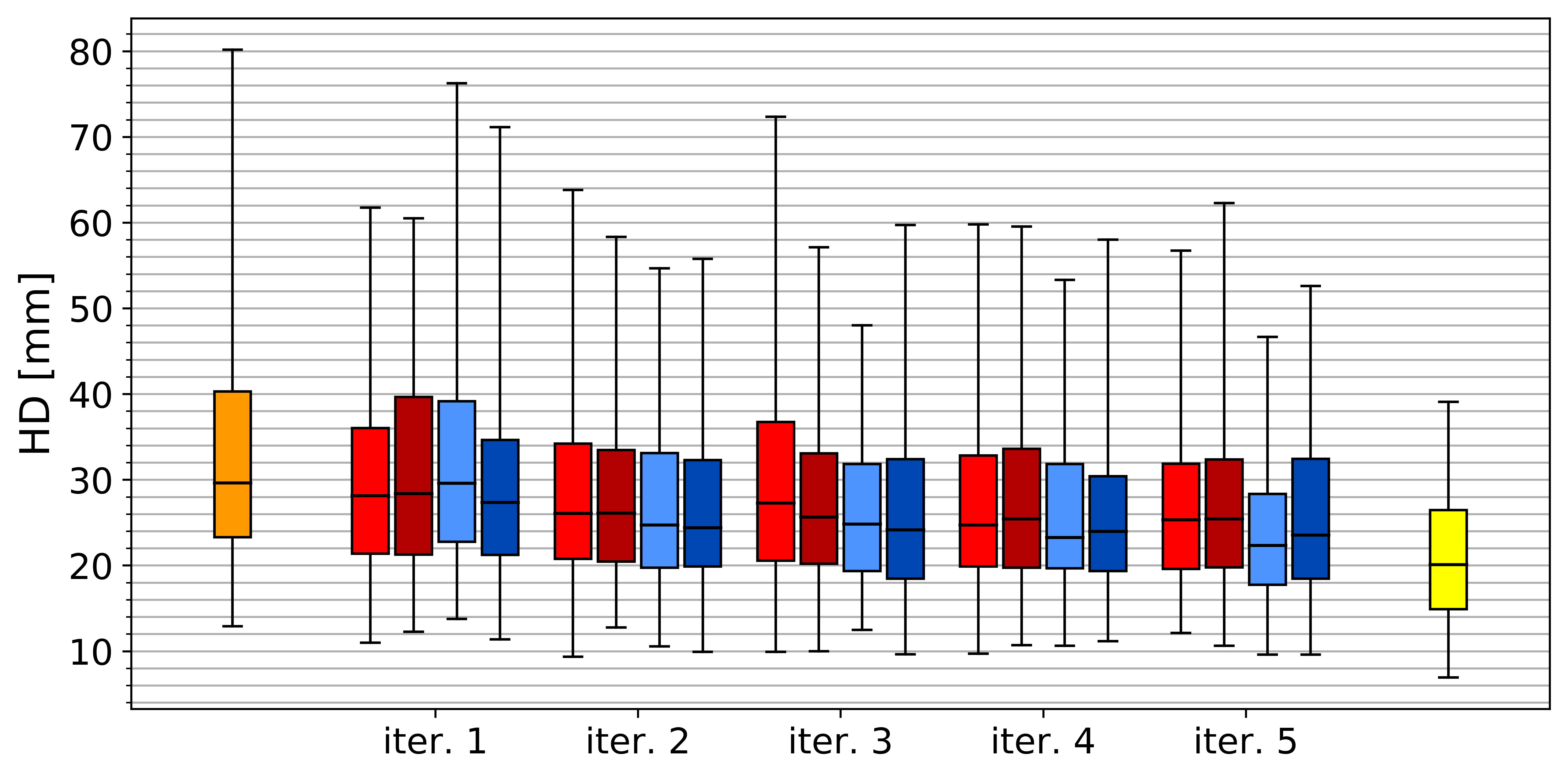}
         \caption{HD (upper cap is 95\textsuperscript{th} percentile)}
         \label{fig:max_dist_iter}
     \end{subfigure}
 \vspace*{2mm}

        \caption{Box plots summarizing evaluation results for models trained throughout five active learning iterations. For reference, results of the initial and whole data pool models are included. }
        \label{fig:iter_results}
\end{figure*}

\begin{figure*}
     \def \wfactor {0.49}
     \centering
     \begin{subfigure}[b]{\wfactor\textwidth}
         \centering
         \includegraphics[width=\textwidth]{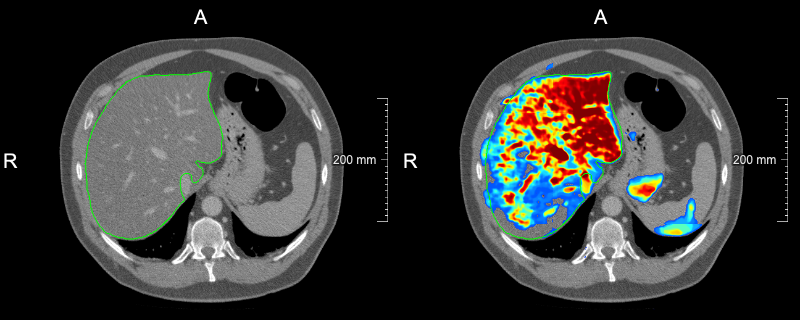}
         \caption{slice-level uncertainty: $0.066$}
     \end{subfigure}
     \hfill
     \begin{subfigure}[b]{\wfactor\textwidth}
         \centering
         \includegraphics[width=\textwidth]{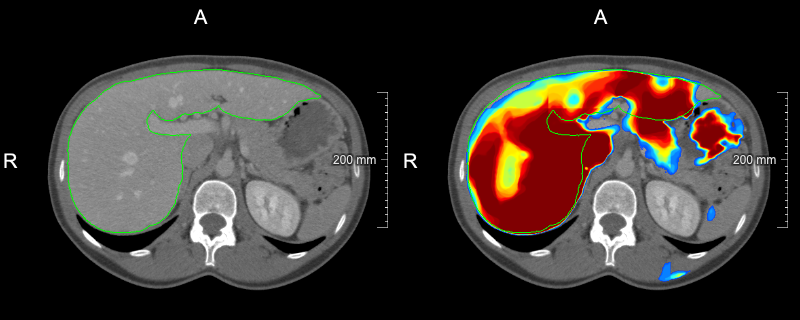}
         \caption{slice-level uncertainty: $0.056$}
     \end{subfigure}
     
     \begin{subfigure}[b]{\wfactor\textwidth}
         \centering
         \includegraphics[width=\textwidth]{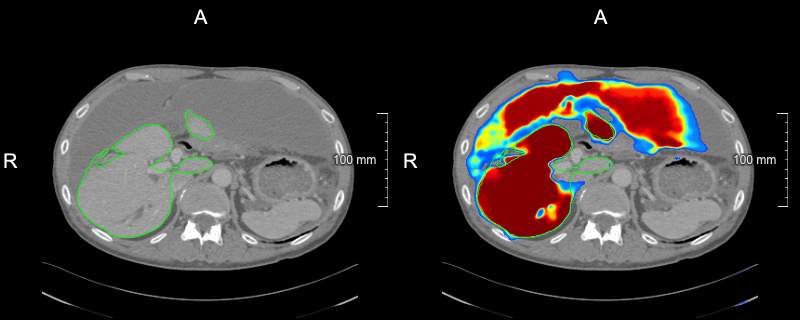}
         \caption{slice-level uncertainty: $0.052$}
     \end{subfigure}
     \hfill
     \begin{subfigure}[b]{\wfactor\textwidth}
         \centering
         \includegraphics[width=\textwidth]{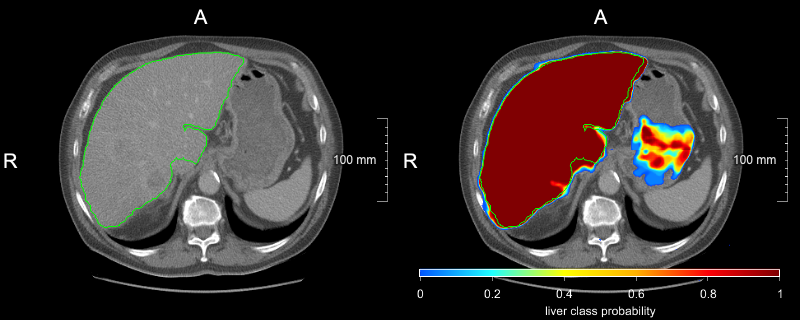}
         \caption{slice-level uncertainty: $0.020$}
     \end{subfigure}
     
    \caption{Examples of slices selected in the first USS iteration with overlaid liver reference segmentation (green contour) and model liver probability output (heatmap): (a)-(c) slices with the biggest slice level uncertainty, (d) slice with the lowest uncertainty among selected ones.}
    \label{fig:uncertain_slq_slices}
\end{figure*}

\begin{figure*}
     \def \wfactor {0.9}
     \centering
     \begin{subfigure}[b]{\wfactor\textwidth}
         \centering
         \includegraphics[width=\textwidth]{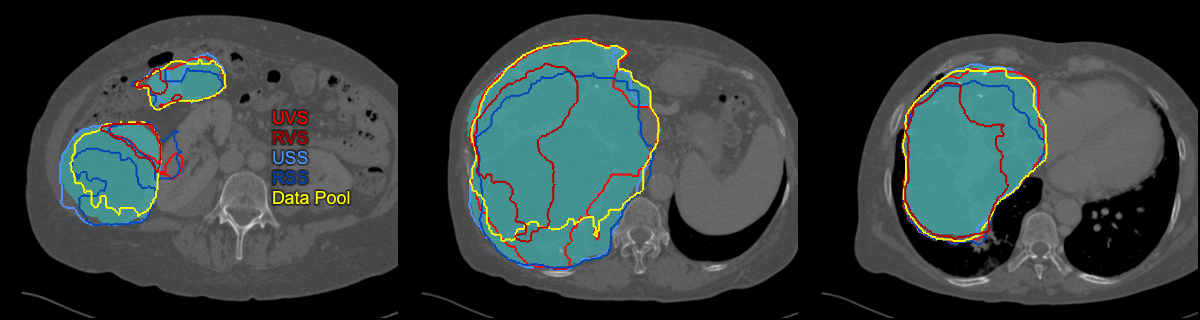}
         \caption{Polycystic liver case where the USS strategy resulted in the best segmentation: \emph{UVS} 0.83 Dice, 25.2\% RVE, 8.0\,mm MSD, 64\,mm HD; \emph{RVS} 0.53, 63.2\%, 18.0\,mm, 64\,mm; \emph{USS} 0.94, 5.2\%, 3.2\,mm, 59\,mm; \emph{RSS} 0.84, 21.7\%, 8.7\,mm, 65\,mm; \emph{Data Pool} 0.85, 24\%, 8.3\,mm, 64\,mm.}
         \label{fig:final_examples_polycistic}
     \end{subfigure}
     
     \begin{subfigure}[b]{\wfactor\textwidth}
         \centering
         \includegraphics[width=\textwidth]{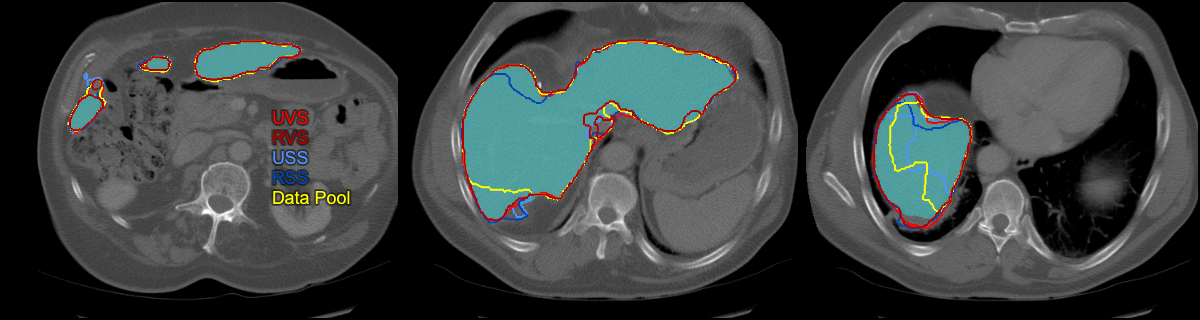}
         \caption{Case where UVS outperformed all models: \emph{UVS} 0.96 Dice, 0.4\% RVE, 1.1\,mm MSD, 17\,mm HD; \emph{RVS} 0.96, 3.1\%, 1.4\,mm, 21\,mm; \emph{USS} 0.94, 5.6\%, 1.5\,mm, 24\,mm; \emph{RSS} 0.95, 2.5\%, 1.4\,mm, 22\,mm; \emph{Data Pool} 0.93, 9.1\%, 2.1\,mm, 32\,mm.}
         \label{fig:final_examples_resected}
     \end{subfigure}
     
     \begin{subfigure}[b]{\wfactor\textwidth}
         \centering
         \includegraphics[width=\textwidth]{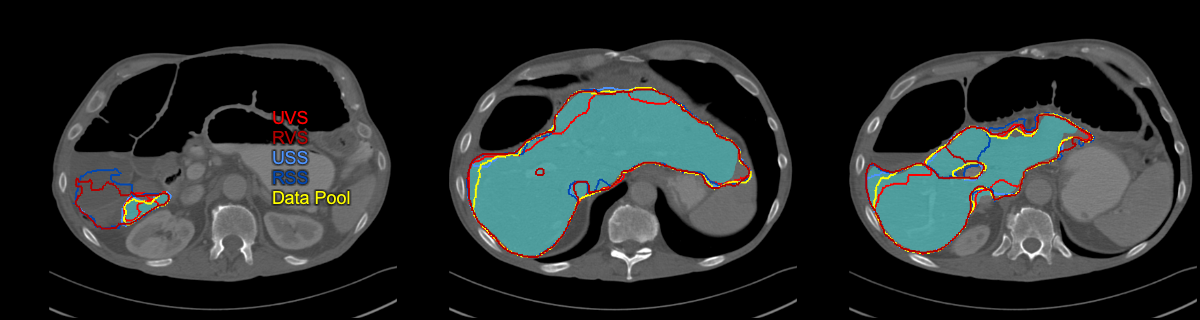}
         \caption{Case where both random strategies resulted in overestimation in the caudal liver region: \emph{UVS} 0.91 Dice, 6.1\% RVE, 2.4\,mm MSD, 23\,mm HD; \emph{RVS} 0.93, 3.4\%, 2.1\,mm, 25\,mm; \emph{USS} 0.95, 1.0\%, 1.5\,mm, 17\,mm; \emph{RSS} 0.77, 41.0\%, 23.3\,mm, 141\,mm; \emph{Data Pool} 0.94, 0.2\%, 1.7\,mm, 21\,mm.}
     \end{subfigure}
     
     \begin{subfigure}[b]{\wfactor\textwidth}
         \centering
         \includegraphics[width=\textwidth]{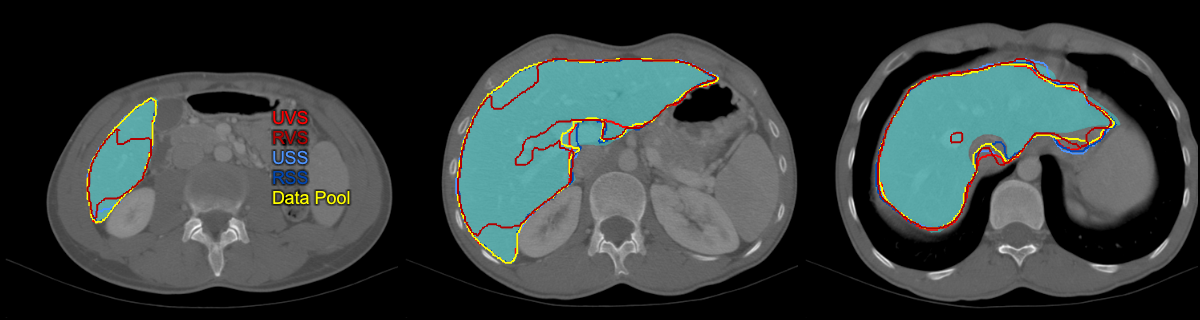}
         \caption{Case for which all strategies except RVS achieved very good segmentation performance: \emph{UVS} 0.97 Dice, 0.1\% RVE, 0.9\,mm MSD, 18\,mm HD; \emph{RVS} 0.91, 11.8\%, 2.9\,mm, 36\,mm; \emph{USS} 0.96, 1.8\%, 1.0\,mm, 20\,mm; \emph{RSS} 0.97, 1.6\%, 0.9\,mm, 20\,mm; \emph{Data Pool} 0.97, 0.0\%, 0.8\,mm, 20\,mm.}
     \end{subfigure}
     
    \caption{Representative examples presenting segmentation output of the converged models and the model trained on the whole data pool.}
    \label{fig:final_examples}
\end{figure*}

\section{Results}
\label{sec:results}

\subsection{Annotation Effort}
The initial model was trained using five fully annotated volumes accounting for a total of 1410\,(501) annotated slices (liver slices).
On average, 1342\,(576) and 1446\,(558) slices were annotated in each UVS and RVS iteration, respectively.
In USS and RSS iterations, we sampled $N_S=200$ ($576/3 \approx 200$) slices.
Consequently, in the last 5\textsuperscript{th} and final iteration, the models were trained using 13\%\,(13\%), 14\%\,(12\%), 4\%\,(5\%), and 6\%\,(6\%) of all slices (liver slices) available in the data pool for the UVS, RVS, USS, and RSS strategy, respectively.
More details can be found in Tab.\,\ref{tab:annotation_stats}.

\begin{table}
\def \colwidth {50pt}
\caption{Training data summary (annotated slices/annotated liver slices/unique patients) over the course of five active learning iteration for each investigated query strategy. }
\setlength{\tabcolsep}{3pt}
\begin{tabular}{l r r r r}
\toprule
 & \multicolumn{1}{c}{UVS} & \multicolumn{1}{c}{RVS} & \multicolumn{1}{c}{USS} & \multicolumn{1}{c}{RSS} \\
\midrule 
initial & \multicolumn{4}{c}{1410/501/5} \\
iter. 1 & 2541/1045/10 & 3206/1070/10 & 1610/669/72 &   1873/701/195 \\
iter. 2 & 3651/1537/15 & 4631/1608/15 & 1810/797/104 &  2323/901/233 \\
iter. 3 & 5105/2224/20 & 5813/2150/20 & 2010/947/125 &  2790/1101/236\\
iter. 4 & 6941/2761/25 & 7094/2687/25 & 2210/1123/149 & 3278/1301/239 \\
iter. 5 & 8121/3382/30 & 8641/3292/30 & 2410/1292/160 & 3730/1501/240 \\
\midrule 
data pool & \multicolumn{4}{c}{60513/26400/242} \\
\bottomrule
\end{tabular}
\label{tab:annotation_stats}
\end{table}


\subsection{Query Strategy Comparison}
For all strategies, an overall improvement in all segmentation metrics was observed over the course of active learning iterations (see Fig.\,\ref{fig:iter_results}).
The mean performance of the initial model was 0.925, 7.87\%, 2.94\,mm, and 36.3\,mm for DICE, RVE, MSD, and HD, respectively.
The mean performance of models resulting from the fifth (last) iteration of the investigated query strategies was in the range of [0.956, 0.960], [3.98, 6.04]\%, [1.4, 1.73]\,mm, and [25.6, 28.6]\,mm for DICE, RVE, MSD, and HD, respectively (see Tab.\,\ref{tab:dice_30epochs}-\ref{tab:hd_30epochs}).
The model obtained from the last iteration of the USS strategy was significantly better (Wilcoxon signed-rank test) than the models from the remaining strategies.
When considering segmentation metrics across all five iterations, the USS strategy resulted in the best-performing model most of the time.
Interestingly, this strategy produced the worst model according to all metrics in the first iteration, even failing to improve the initial model's performance for RVE and HD. 
None of the models trained with 4\%-14\% of the data was able to match the results of the model trained on the whole data pool that achieved on average 0.967 DICE, 3.8\% RVE, 1.18\,mm MSD, and 22.9\,mm HD.

For volume sampling strategies there is no clear difference between uncertainty-based and random sampling.
In the case of slice sampling strategies, the uncertainty-based sampling resulted in a consistent improvement (excluding the first iteration) across all metrics.

\begin{table}
\centering
\def \colwidth {30pt}
\caption{Training step count for models used for evaluation obtained using four investigated query strategies over five active learning iterations. 
For reference, data for the initial and data pool model is given.}
\setlength{\tabcolsep}{3pt}
\begin{tabular}{l r r r r}
\toprule
 & \multicolumn{1}{c}{UVS} & \multicolumn{1}{c}{RVS} & \multicolumn{1}{c}{USS} & \multicolumn{1}{c}{RSS} \\
\midrule 
initial & \multicolumn{4}{c}{12 000} \\
iter. 1 & 35 000 & 37 000 & 40 000 & 46 000 \\
iter. 2 & 61 000 & 64 000 & 61 000 & 67 000 \\
iter. 3 & 80 000 & 83 000 & 71 000 & 94 000 \\
iter. 4 & 75 000 & 85 000 & 91 000 & 117 000 \\
iter. 5 & 119 000 & 98 000 & 98 000 & 139 000 \\
iter. 5 (converged) & 356 000 & 297 000 & 291 000 & 465 000 \\
\midrule 
data pool & \multicolumn{4}{c}{825 000} \\
\bottomrule
\end{tabular}
\label{tab:train_iter}
\end{table}

\begin{figure*}
     \centering
     \begin{subfigure}[b]{0.45\textwidth}
         \centering
         \includegraphics[width=\textwidth]{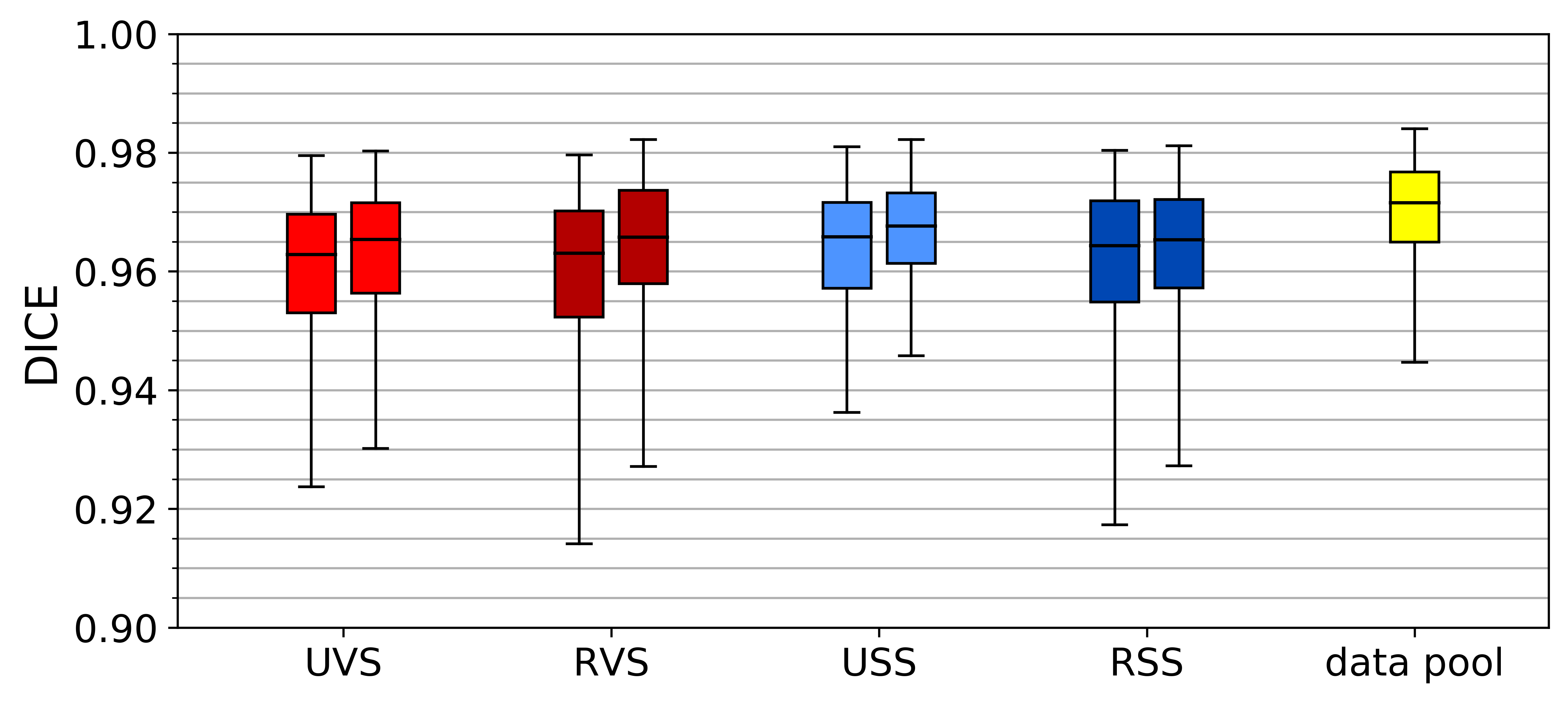}
         \caption{DICE (lower cap is 5\textsuperscript{th} percentile)}
         \label{fig:dice_converged}
     \end{subfigure}
     \hfill
     \begin{subfigure}[b]{0.45\textwidth}
         \centering
         \includegraphics[width=\textwidth]{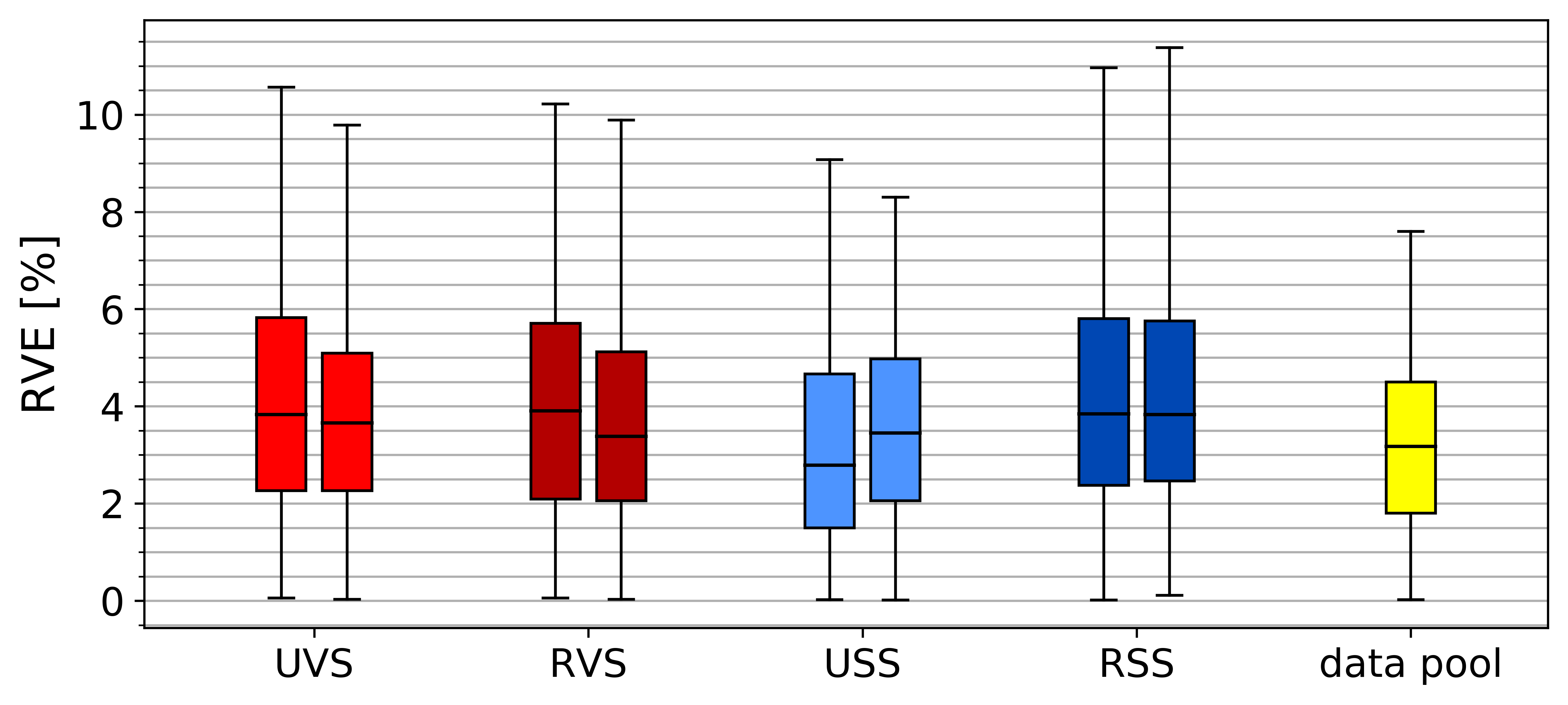}
         \caption{RVE (upper cap is 95\textsuperscript{th} percentile)}
         \label{fig:rve_converged}
     \end{subfigure}
     \begin{subfigure}[b]{0.45\textwidth}
         \centering
         \includegraphics[width=\textwidth]{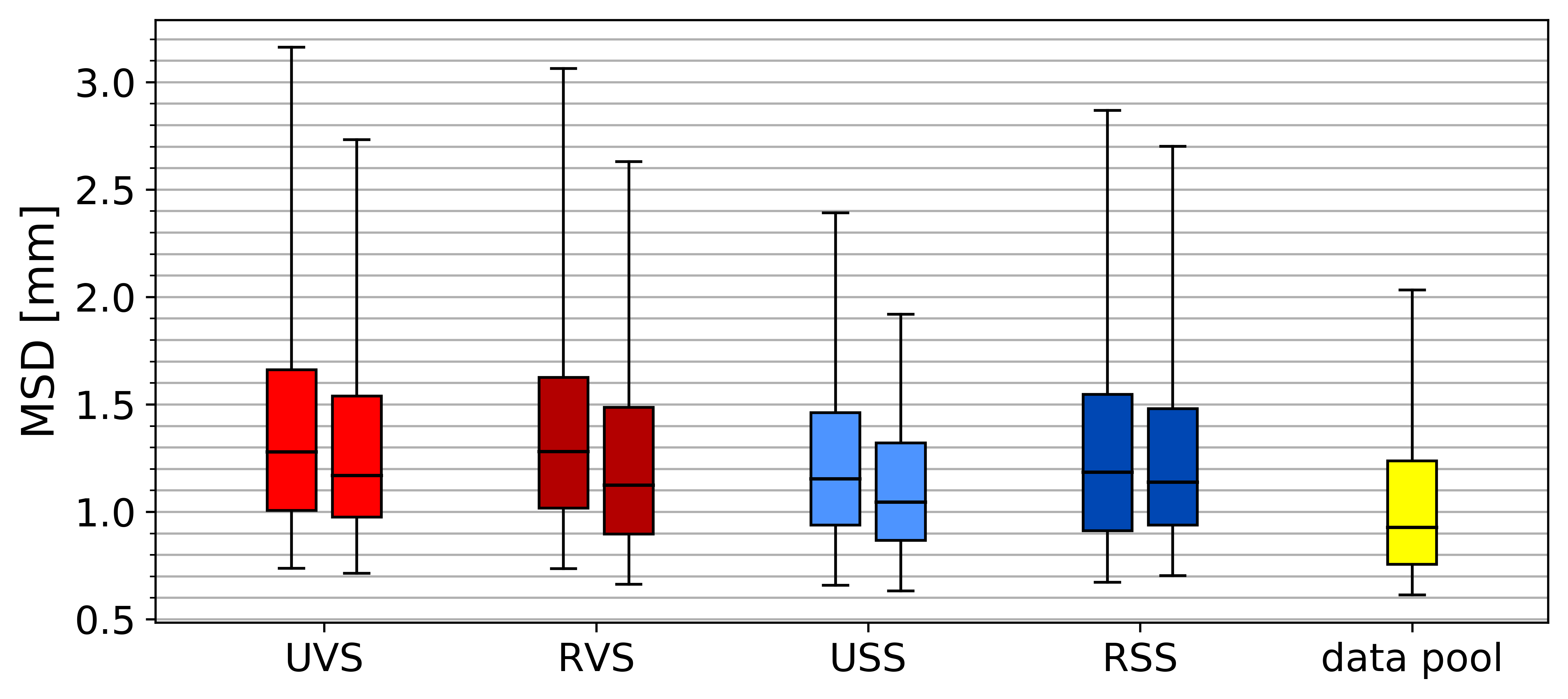}
         \caption{MSD (upper cap is 95\textsuperscript{th} percentile)}
         \label{fig:mean_dist_converged}
     \end{subfigure}
     \hfill
     \begin{subfigure}[b]{0.45\textwidth}
         \centering
         \includegraphics[width=\textwidth]{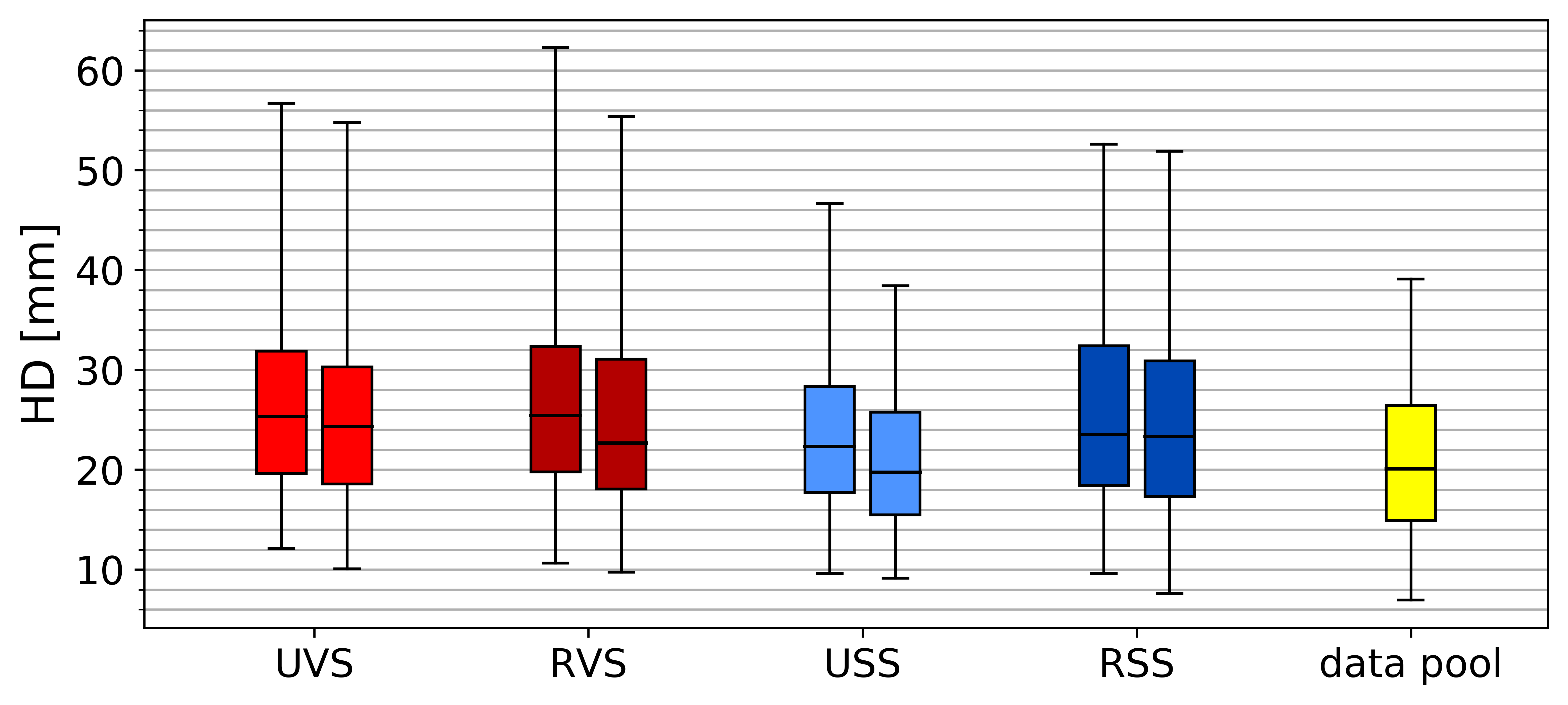}
         \caption{HD (upper cap is 95\textsuperscript{th} percentile)}
         \label{fig:max_dist_converged}
     \end{subfigure}
        \caption{Box plots summarizing evaluation results for models trained for max of 30 epochs (left) and until convergence (right) using data from the fifth iteration. For reference, results of the whole data pool model are included. }
        \label{fig:converged_results}
\end{figure*}

\subsection{Converged Models Comparison}
Training until convergence using data resulting from the fifth active learning iteration resulted in an improvement of all segmentation metrics across all query strategies with one exception of RVE metric for the USS strategy (see Tab.\,\ref{tab:results_converged} and  Fig.\,\ref{fig:converged_results}).
The model trained on data resulting from the RSS strategy required the longest training time (465k iterations), whereas the USS model needed the fewest iterations (291k) compared to 825k iterations for the model trained on the whole data pool.
Among the investigated strategies, the USS model had the best performance when compared to other models with most of the differences being significant.
None of the models achieved comparable results to the whole data pool model, but we observed that the USS model resulted in more robust segmentations than the whole data pool model according to the 5\textsuperscript{th} percentile for DICE and the 95\textsuperscript{th} percentile for MSD and HD metric.

To quantify the effect of uncertainty-based vs random sampling and slice vs volume sampling on the 5\textsuperscript{th} percentile of DICE and 95\textsuperscript{th} percentile of the remaining metrics we performed a quantile regression analysis using evaluation results of the converged models (Tab.\,\ref{tab:regression}).
Uncertainty sampling resulted in a significant improvement w.r.t. random sampling equal to 0.012 for the 5\textsuperscript{th} percentile of DICE and -0.41 for the 95\textsuperscript{th} percentile of MSD.
Similarly, changing from volume to slice sampling results in an increase of 0.012 in the 5\textsuperscript{th} percentile DICE and a decrease of 0.48 in the 95\textsuperscript{th} percentile of MSD.
The effect on the remaining metrics did not pass our significance test (95\% confidence interval included 0).

\begin{table*}
\centering
\def \colwidth {100pt}
\caption{DICE for models trained using the investigated query strategies over five active learning iterations. For reference, results of the initial model and the model trained on the whole data pool are given.}
\setlength{\tabcolsep}{3pt}
\begin{tabular}{lM{\colwidth}M{\colwidth}M{\colwidth}M{\colwidth}}
\toprule
 & UVS & RVS & USS & RSS \\
\midrule 
initial & \multicolumn{4}{c}{$0.925\pm0.068\,(0.837)$} \\
iter. 1 & $0.942\pm0.037\,(0.893)$ & $0.937\pm0.038\,(0.868)^{***}$ & $0.932\pm0.048\,(0.866)^{***}$ & $\mathbf{0.943\pm0.041\,(0.887)}$ \\
iter. 2 & $0.947\pm0.033\,(0.895)^{***}$ & $0.947\pm0.034\,(0.903)^{***}$ & $\mathbf{0.951\pm0.044\,(0.914)}$ & $0.948\pm0.042\,(0.899)^{***}$ \\
iter. 3 & $0.945\pm0.036\,(0.904)^{***}$ & $\mathbf{0.955\pm0.028\,(0.917)}$ & $0.953\pm0.028\,(0.919)^{***}$ & $0.951\pm0.047\,(0.912)$ \\
iter. 4 & $0.952\pm0.028\,(0.916)^{***}$ & $0.952\pm0.033\,(0.915)^{***}$ & $\mathbf{0.958\pm0.028\,(0.932)}$ & $0.957\pm0.027\,(0.919)$ \\
iter. 5 & $0.956\pm0.028\,(0.924)^{***}$ & $0.956\pm0.029\,(0.914)^{***}$ & $\mathbf{0.960\pm0.027\,(0.936)}$ & $0.956\pm0.042\,(0.917)^{***}$ \\
\midrule 
data pool & \multicolumn{4}{c}{$0.967\pm0.023\,(0.945)$} \\
\bottomrule
\multicolumn{5}{l}{Best result in a row according to the mean is indicated in bold. Value in parentheses denotes the 5\textsuperscript{th} percentile.} \\
\multicolumn{5}{l}{${}^*p<0.05$, ${}^{**}p<0.01$, ${}^{***}p<0.001$.}
\end{tabular}
\label{tab:dice_30epochs}
\end{table*}

\begin{table*}
\centering
\def \colwidth {100pt}
\caption{RVE for models trained using the investigated query strategies over five active learning iterations.}
\setlength{\tabcolsep}{3pt}
\begin{tabular}{lM{\colwidth}M{\colwidth}M{\colwidth}M{\colwidth}}
\toprule
 & UVS & RVS & USS & RSS \\
\midrule
initial & \multicolumn{4}{c}{$7.87\pm10.0\,(23.32)$} \\
iter. 1 & $\mathbf{5.58\pm8.95\,(13.66)}$ & $6.94\pm5.91\,(17.64)^{***}$ & $11.62\pm16.1\,(26.97)^{***}$ & $8.05\pm12.18\,(19.96)^{***}$ \\
iter. 2 & $\mathbf{5.1\pm6.04\,(13.69)}$ & $6.15\pm6.73\,(13.57)^{***}$ & $5.51\pm15.96\,(10.63)^{**}$ & $7.7\pm13.25\,(18.74)^{***}$ \\
iter. 3 & $7.14\pm8.94\,(16.88)^{***}$ & $\mathbf{4.19\pm5.13\,(11.08)}$ & $6.16\pm6.5\,(14.8)^{***}$ & $7.32\pm17.1\,(14.89)^{***}$ \\
iter. 4 & $6.06\pm6.93\,(12.44)^{***}$ & $\mathbf{5.12\pm5.11\,(11.79)}$ & $5.19\pm7.81\,(10.17)$ & $5.25\pm6.66\,(11.11)$ \\
iter. 5 & $5.1\pm6.84\,(10.56)^{***}$ & $4.81\pm5.25\,(10.22)^{***}$ & $\mathbf{3.98\pm6.97\,(9.07)}$ & $6.04\pm16.23\,(10.97)^{***}$ \\\midrule
data pool & \multicolumn{4}{c}{$3.8\pm4.96\,(7.6)$} \\

\bottomrule
\multicolumn{5}{l}{Best result in a row according to the mean is indicated in bold. Value in parentheses denotes the 95\textsuperscript{th} percentile.} \\
\multicolumn{5}{l}{${}^*p<0.05$, ${}^{**}p<0.01$, ${}^{***}p<0.001$.}
\end{tabular}
\label{tab:rve_30epochs}
\end{table*}

\begin{table*}
\centering
\def \colwidth {100pt}
\caption{MSD for models trained using the investigated query strategies over five active learning iterations.}
\setlength{\tabcolsep}{3pt}
\begin{tabular}{lM{\colwidth}M{\colwidth}M{\colwidth}M{\colwidth}}
\toprule
 & UVS & RVS & USS & RSS \\
\midrule
initial & \multicolumn{4}{c}{$2.94\pm3.9\,(6.94)$} \\
iter. 1 & $\mathbf{2.1\pm1.9\,(4.06)}$ & $2.17\pm1.44\,(4.88)^{**}$ & $2.81\pm4.43\,(6.41)^{***}$ & $2.28\pm3.52\,(4.84)$ \\
iter. 2 & $\mathbf{1.85\pm1.29\,(3.56)}$ & $1.86\pm1.81\,(3.6)$ & $1.96\pm4.89\,(3.04)^{***}$ & $2.04\pm4.3\,(3.5)^{***}$ \\
iter. 3 & $2.12\pm2.45\,(3.96)^{***}$ & $\mathbf{1.64\pm1.5\,(3.11)}$ & $1.64\pm1.26\,(3.2)^{**}$ & $2.05\pm4.87\,(3.0)$ \\
iter. 4 & $1.69\pm1.15\,(2.94)^{***}$ & $1.68\pm1.27\,(3.09)^{***}$ & $\mathbf{1.52\pm1.66\,(2.48)}$ & $1.58\pm1.55\,(2.75)$ \\
iter. 5 & $1.59\pm1.27\,(3.16)^{***}$ & $1.58\pm1.26\,(3.06)^{***}$ & $\mathbf{1.4\pm1.44\,(2.39)}$ & $1.73\pm4.23\,(2.87)^{**}$ \\\midrule
data pool & \multicolumn{4}{c}{$1.18\pm1.1\,(2.03)$} \\

\bottomrule
\multicolumn{5}{l}{Best result in a row according to the mean is indicated in bold. Value in parentheses denotes the 95\textsuperscript{th} percentile.} \\
\multicolumn{5}{l}{${}^*p<0.05$, ${}^{**}p<0.01$, ${}^{***}p<0.001$.}
\end{tabular}
\label{tab:msd_30epochs}
\end{table*}

\begin{table*}
\centering
\def \colwidth {100pt}
\caption{HD for models trained using the investigated query strategies over five active learning iterations.}
\setlength{\tabcolsep}{3pt}
\begin{tabular}{lM{\colwidth}M{\colwidth}M{\colwidth}M{\colwidth}}
\toprule
 & UVS & RVS & USS & RSS \\
\midrule
initial & \multicolumn{4}{c}{$36.3\pm20.9\,(80.2)$} \\
iter. 1 & $\mathbf{32.1\pm16.9\,(61.8)}$ & $32.1\pm15.3\,(60.5)$ & $36.4\pm26.0\,(76.3)^{*}$ & $33.2\pm23.5\,(71.1)$ \\
iter. 2 & $29.9\pm14.5\,(63.8)$ & $29.9\pm15.6\,(58.3)$ & $\mathbf{29.6\pm21.4\,(54.6)}$ & $29.6\pm22.0\,(55.8)$ \\
iter. 3 & $32.4\pm20.1\,(72.3)^{***}$ & $29.5\pm17.2\,(57.1)$ & $\mathbf{27.5\pm12.8\,(48.0)}$ & $29.5\pm24.1\,(59.7)$ \\
iter. 4 & $28.3\pm13.3\,(59.8)$ & $28.7\pm13.5\,(59.5)$ & $\mathbf{27.4\pm13.9\,(53.3)}$ & $27.9\pm16.5\,(58.0)$ \\
iter. 5 & $28.5\pm13.8\,(56.7)^{***}$ & $28.6\pm14.0\,(62.3)^{***}$ & $\mathbf{25.6\pm15.0\,(46.6)}$ & $28.2\pm20.9\,(52.6)^{***}$ \\\midrule
data pool & \multicolumn{4}{c}{$22.9\pm12.7\,(39.1)$} \\

\bottomrule
\multicolumn{5}{l}{Best result in a row according to the mean is indicated in bold. Value in parentheses denotes the 95\textsuperscript{th} percentile.} \\
\multicolumn{5}{l}{${}^*p<0.05$, ${}^{**}p<0.01$, ${}^{***}p<0.001$.}
\end{tabular}
\label{tab:hd_30epochs}
\end{table*}

\begin{table*}
\def \colwidth {90pt}
\centering
\caption{Evaluation results for the converged models.}
\setlength{\tabcolsep}{3pt}
\begin{tabular}{M{30pt}M{\colwidth}M{\colwidth}M{\colwidth}M{\colwidth}}
\toprule
 & DICE & RVE\,[\%] & MSD\,[mm] & HD\,[mm] \\
\midrule 
UVS & $0.959\pm0.027\,(0.930)^{***}$ & $4.67\pm6.23\,(9.78)^{*}$ & $1.45\pm1.18\,(2.73)^{***}$ & $27.0\pm12.4\,(54.8)^{***}$ \\
RVS & $0.959\pm0.035\,(0.927)^{***}$ & $4.41\pm5.33\,(9.90)$ & $1.44\pm1.42\,(2.63)^{***}$ & $26.4\pm13.2\,(55.4)^{***}$ \\
USS & $\mathbf{0.964\pm0.025\,(0.946)}$ & $\mathbf{4.20\pm6.63\,(8.30)}$ & $\mathbf{1.35\pm2.94\,(1.92)}$ & $\mathbf{23.4\pm19.0\,(38.4)}$ \\
RSS & $0.959\pm0.029\,(0.927)^{***}$ & $5.00\pm6.44\,(11.38)^{***}$ & $1.54\pm1.94\,(2.70)^{***}$ & $27.1\pm16.1\,(52.0)^{***}$ \\
\midrule
data pool & $0.967\pm0.023\,(0.945)$ & $3.80\pm4.96\,(7.60)$ & $1.18\pm1.10\,(2.03)$ & $22.9\pm12.7\,(39.1)$ \\
\bottomrule
\multicolumn{5}{l}{Best result for given metric according to the mean is indicated in bold (model trained on the whole data pool is excluded from comparison).} \\
\multicolumn{5}{l}{${}^*p<0.05$, ${}^{**}p<0.01$, ${}^{***}p<0.001$.}
\end{tabular}
\label{tab:results_converged}
\end{table*}

\begin{table*}
\def \colwidth {90pt}
\centering
\caption{Quantile regression results quantifying the effect of uncertainty vs random sampling and slice vs volume sampling on the 5\textsuperscript{th} percentile of DICE and 95\textsuperscript{th} percentile of RVE, MSD, and HD. The analysis was run using evaluation results of the converged models.}
\setlength{\tabcolsep}{3pt}
\begin{tabular}{l M{65pt}M{25pt} M{65pt}M{24pt} M{63pt}M{24pt} M{60pt}M{24pt}}
\toprule
& \multicolumn{2}{c}{DICE\dag} & \multicolumn{2}{c}{RVE\,[\%]\ddag} & \multicolumn{2}{c}{MSD\,[mm]\ddag} & \multicolumn{2}{c}{HD\,[mm]\ddag}\\
 & coefficient & p-value & coefficient & p-value & coefficient & p-value & coefficient & p-value \\
\midrule 
Intercept & 0.920 [0.914, 0.925] & <\,0.001 & 11.21 [9.26, 13.16] & <\,0.001 & 2.90 [2.64, 3.16] & <\,0.001 & 59.6 [50.0, 69.3] & <\,0.001 \\
Uncertainty Sampling & 0.012 [0.005, 0.019] & <\,0.001 & -1.90 [-4.21, 0.42] & 0.1 & -0.41 [-0.73, -0.08] & 0.015 & -4.8 [-16.3, 6.6] & 0.4 \\
Slice Sampling & 0.012 [0.005, 0.019] & <\,0.001 & -0.80 [-3.12, 1.51] & 0.5 & -0.48 [-0.81, -0.16] & 0.004 & -7.7 [-19.1, 3.7] & 0.2 \\
\bottomrule
\multicolumn{9}{l}{Results from 0.05(\dag), 0.95(\ddag) quantile regression. Values in brackets denote 95\% confidence interval.}
\end{tabular}
\label{tab:regression}
\end{table*}

\section{Discussion}
\label{sec:discussion}
In our work, we proposed the uncertainty slice sampling (USS) strategy in the context of pool-based active learning.
Our strategy selects 2D image slices from a pool of 3D volumes using aggregated voxel-wise predictive entropy as the uncertainty measure.
We evaluated the proposed strategy on a CT liver segmentation task and compared it with random slice sampling (RSS), uncertainty volume sampling (UVS), and random volume sampling (RVS) strategies.
The model trained using the USS data (4\% of available data) achieved significantly better results than the remaining strategies.
Although after five active learning iterations the USS model was inferior in performance on average to the model trained on all available data, it provided more robust segmentation as measured by 5\textsuperscript{th} DICE and 95\textsuperscript{th} MSD metrics.
We hypothesize that this can be attributed to differences in the training set composition.
The training set resulting from the USS contains a bigger proportion of difficult/rare cases compared to the whole data-pool training set, which effectively causes that the model sees them more frequently during the training process.
Fig.\,\ref{fig:final_examples} shows exemplary outputs from the investigated models including two hard cases from the test set: a polycystic (Fig.\,\ref{fig:final_examples_polycistic}) and a resected (Fig.\,\ref{fig:final_examples_resected}) liver.
We think that the robustness of the whole data pool model could be increased by employing a hard example mining during training to dynamically adjust the sampling rate of difficult examples\cite{shrivastava2016training, bian2018pyramid}.
Selecting only uncertain cases in the course of active learning can overload the model with difficult examples causing a performance drop.
This can be observed for the USS strategy after the first iteration (see Fig.\,\ref{fig:iter_results}, where the model performs substantially worse than its random counterpart (RSS).

As our strategy relies on the model's uncertainty to query cases, the confidence calibration of a model can have a substantial impact on which cases are deemed uncertain. 
Recently, it has been shown that modern deep neural networks do not output well-calibrated probabilities and tend to be overconfident\cite{guo2017calibration}.
In our work, we have used MC dropout that improves the calibration quality of models trained with the Dice loss\cite{mehrtash2020confidence}.
Exemplary probability maps produced by our model are shown in Fig.\,\ref{fig:uncertain_slq_slices}.
We think that investigation of various calibration techniques, e.g., deep ensembles and temperature scaling, in the context of active learning could be an interesting future research direction.

In our study, we focused on a detailed investigation of the proposed query strategy and we didn't intend to achieve state-of-the-art results on the CT liver segmentation task.
We are aware that in addition to using more training data, increasing model capacity or changing the architecture could further boost the segmentation performance.
The influence of various neural network designs on the efficiency of our proposed query strategy would in our opinion make a compelling experiment.
Moreover, we think that investigation of active learning together with AutoML frameworks such as nnU-net\cite{isensee2021nnu}, which make state-of-the-art segmentation accessible to people without ML expertise, would be an interesting extension of our work.



\noindent
\textbf{Limitations} Our conclusions are based on the assumption w.r.t. the annotation effort derived from empirical observations of annotation workflows. 
These assumptions might not hold for some annotators or if different labeling tools are used.

\section*{Acknowledgment}
We would like to thank our clinical partners from Yokohama City University, Yokohama, Japan, St\"{a}dtisches Klinikum Dresden, Dresden, Germany, and Radboud University Clinical Center, Nijmegen, the Netherlands for providing imaging data used in this study.
We acknowledge the organizers of the LiTS and the CHAOS competitions for making the training data publicly available.
We are thankful to Christiane Engel and Andrea Koller for providing manual segmentations.
The work was funded by the Fraunhofer-Gesellschaft.


\bibliographystyle{unsrt}
\bibliography{bibliography}

\EOD

\end{document}